\documentclass{article}
\pdfoutput=1

\usepackage{PRIMEarxiv}

\usepackage[utf8]{inputenc} 
\usepackage[T1]{fontenc}    
\usepackage{hyperref}       
\usepackage{url}            
\usepackage{booktabs}       
\usepackage{amsfonts}       
\usepackage{nicefrac}       
\usepackage{microtype}      
\usepackage{lipsum}
\usepackage{fancyhdr}       
\usepackage{graphicx}       
\graphicspath{{media/}}     
\usepackage{subfig}
\usepackage{array, caption, threeparttable}
\usepackage{multirow}
\usepackage{amsmath}
\title{Doctor Sun: A Bilingual Multimodal Large Language Model for Biomedical AI
}

\author{
  Dong Xue, Ziyao Shao, Zhaoyang Duan \\
  Key Laboratory of Smart Manufacturing in Energy Chemical Process, Ministry of Education \\
  East China University of Science and Technology \\
  Shanghai 200237, China\\
  \texttt{dong.xue@ecust.edu.cn,y30230950@mail.ecust.edu.cn,zhaoyangduan@ecust.edu.cn} \\
   \And
  Fangzhou Liu \\
  Research Institute of Intelligent Control and Systems \\
  Harbin Institute of Technology \\
  Harbin 150001, China\\
  \texttt{fangzhou.liu@hit.edu.cn} \\
   \And
  Bing Li \\
  Department of Emergency Medicine, Sir Run Run Shaw Hospital \\
  Zhejiang University School of Medicine \\
  Hangzhou 310000, China\\
  \texttt{sjtuspring@163.com} \\
  \And
  Zhongheng Zhang \\
  Provincial Key Laboratory of Precise Diagnosis \\
  Treatment of Abdominal Infection, Sir Run Run Shaw Hospital \\
  Zhejiang University School of Medicine\\
  Hangzhou 310000, China\\
  School of Medicine \\
  Shaoxing University\\
  Shaoxing 311800, China\\
  \texttt{zh\_zhang1984}@zju.edu.cn} 

\begin{document}
\maketitle

\begin{abstract}
Large multimodal models (LMMs) have demonstrated significant potential in providing innovative solutions for various biomedical tasks, including pathology analysis, radiology report generation, and biomedical assistance. However, the existing multimodal biomedical AI is typically based on foundation LLMs, thus hindering the understanding of intricate medical concepts with limited medical training data. Moreover, recent LLaVA-induced medical LMMs struggle to effectively capture the intricate relationship between the texts and the images. Therefore, we introduce Doctor Sun, a large multimodal generative model specialized in medicine, developed to encode, integrate, and interpret diverse biomedical data modalities such as text and images. In particular, Doctor Sun integrates a pre-trained vision encoder with a medical LLM and conducts two-stage training on various medical datasets, focusing on feature alignment and instruction tuning. Moreover, we release SunMed-VL, a wide-range bilingual medical multimodal dataset, along with all associated models, code, and resources, to freely support the advancement of biomedical multimodal research. 
\end{abstract}

\keywords{Multimodal Machine Learning, Large Language Model, Medical Diagnosis }

\flushbottom
\maketitle
%
%
\thispagestyle{empty}


\section*{Introduction}
In recent years, large language models (LLMs) have demonstrated remarkable capabilities in rapidly acquiring skills by integrating past experiences and external data sources. This proficiency is particularly valuable in the medical field, where the extensive knowledge base of experienced physicians often leads to superior diagnostic and treatment outcomes. By emulating this knowledge acquisition and application process, LLMs have the potential to significantly enhance medical practice and decision-making significantly, thereby reducing diagnostic errors and improving patient care outcomes. However, existing LLMs primarily rely on textual data, which limits their ability to fully capture the complexity of medical diagnostics that often requires multimodal data integration. For instance, many critical diagnostic indicators are found in medical images, such as radiological scans or pathological slides, which provide essential anatomical and physiological insights. Recent studies~\cite{xiong2023doctorglm, Wang2023HuaTuoTL} have introduced various medical LLMs, pioneering new avenues in computational pathology and precision medicine, but their reliance on text-only data highlights the need for models that can effectively integrate multimodal information to address the challenges of real-world medical scenarios.

Despite the impressive diagnostic abilities of medical LLMs, relying solely on textual data for medical diagnosis is both conservative and inadequate. Many critical diagnostic indicators are found in medical images, such as radiological scans, pathological slides, and clinical photographs, which provide essential anatomical and physiological insights~\cite{hussain2022modern}. For example, hippocampal volume is a well-validated sMRI feature for classifying Alzheimer's disease, while chest X-rays play a crucial role in diagnosing lung diseases. Therefore, integrating multimodal data into these models is indispensable~\cite{fahad2025advancements}. By combining text-based clinical data with medical images, multimodal large language models (MLLMs) can capture a broader spectrum of disease manifestations, enabling a more nuanced understanding of complex conditions. However, achieving effective integration of multimodal data remains challenging due to issues such as feature alignment, limited domain-specific training data, and the need for models to balance general-purpose capabilities with medical expertise~\cite{wu2023towards}.

Previous research on medical MLLMs often restricts models to single-language interactions and focuses on diagnosing specific illnesses. For instance, XrayGLM~\cite{wang2023XrayGLM} is primarily designed for lung disease diagnosis and is limited to English-speaking contexts. However, real-world medical practice involves addressing diverse and complex disease patterns, which require extensive medical knowledge across multiple disciplines, such as geriatric medicine, internal medicine, and palliative care. For example, developing treatment plans for elderly patients with chronic conditions demands a nuanced understanding of comorbidities, patient-specific factors, and multidisciplinary expertise~\cite{zhang2023biomedgpt}. While certain multimodal models, such as LLaVA-Med~\cite{li2024llava}, have demonstrated the ability to handle a broad range of diagnostic tasks, their performance on complex and interdisciplinary medical challenges remains suboptimal. This limitation stems from the insufficient medical knowledge embedded in their foundational models, which are often adapted from general-purpose LLMs with limited domain-specific training. Furthermore, these models struggle to bridge the gap between generic and medical knowledge, as their general base models lack the depth of expertise required for advanced diagnostic reasoning and treatment planning. Therefore, there is a pressing need for multimodal models that not only integrate extensive medical knowledge but also excel in handling complex diagnostic and therapeutic scenarios while maintaining general-purpose capabilities.

In this article, we introduce Doctor Sun, a bilingual (Chinese-English) MLLM specifically designed to advance medical diagnostics across multiple specialties. Doctor Sun integrates three key components: a text foundation model for logical reasoning and clinical decision-making, a visual foundation model to extract image features and identify abnormalities in medical scans, and a cross-modal projector to align and map visual data into the textual semantic space. This architecture enables the seamless integration of imaging findings with clinical notes, providing a comprehensive understanding of patient conditions. The model is trained on a meticulously curated, high-quality bilingual dataset derived from public sources, encompassing radiology images, pathology slides, and clinical photographs, along with corresponding textual annotations in both Chinese and English. To ensure domain-specific expertise, the general-purpose language foundation model of Doctor Sun is first pre-trained and optimized to accumulate fundamental medical knowledge. Subsequently, the entire model undergoes a two-stage training strategy, focusing on feature alignment and instruction tuning, to achieve proficiency in multimodal medical diagnostic tasks while retaining general-purpose capabilities. In summary, this article presents several key contributions:
\begin{itemize}
    \item By implementing medical domain-specific offsets for vision encoders and fine-tuning the language foundation model within the medical domain, Doctor Sun accumulates extensive pathology knowledge and achieves a nuanced understanding of pathological images and texts, enabling it to perform complex pathological diagnoses. Empirical studies demonstrate the superior performance of Doctor Sun, achieving excellent zero-shot evaluation results.
    \item The optimal ratio of domain-specific to generic data in feature alignment and instruction tuning is investigated, enhancing Doctor Sun's proficiency in medical diagnostics while maintaining its general-purpose capabilities.
    \item A selective mixed Chinese-English medical visual-language dataset, SunMed-VL, is developed, spanning multiple medical specialties such as respiratory, oncology, neurology, and orthopedics, ensuring broad and diverse coverage of medical domains. All model codes, datasets, and modeling weights are publicly available.
\end{itemize}

The remainder of this article is organized as follows: Section 2 introduces the data processing, model architecture, and training methods of Doctor Sun, as well as the establishment process of the medical language foundation model. Section 3 presents the experimental results, accompanied by a thorough analysis across various tasks and datasets, as well as comparisons with state-of-the-art benchmarks. Section 4 discusses the experimental findings and conducts additional experiments on the medical language foundation model to analyze its performance and the effectiveness of the training method. Finally, Section 5 concludes the study by summarizing the key findings.

\section*{Materials and methods}
\begin{figure}[t]
\centering
{\includegraphics[width=0.95\linewidth]{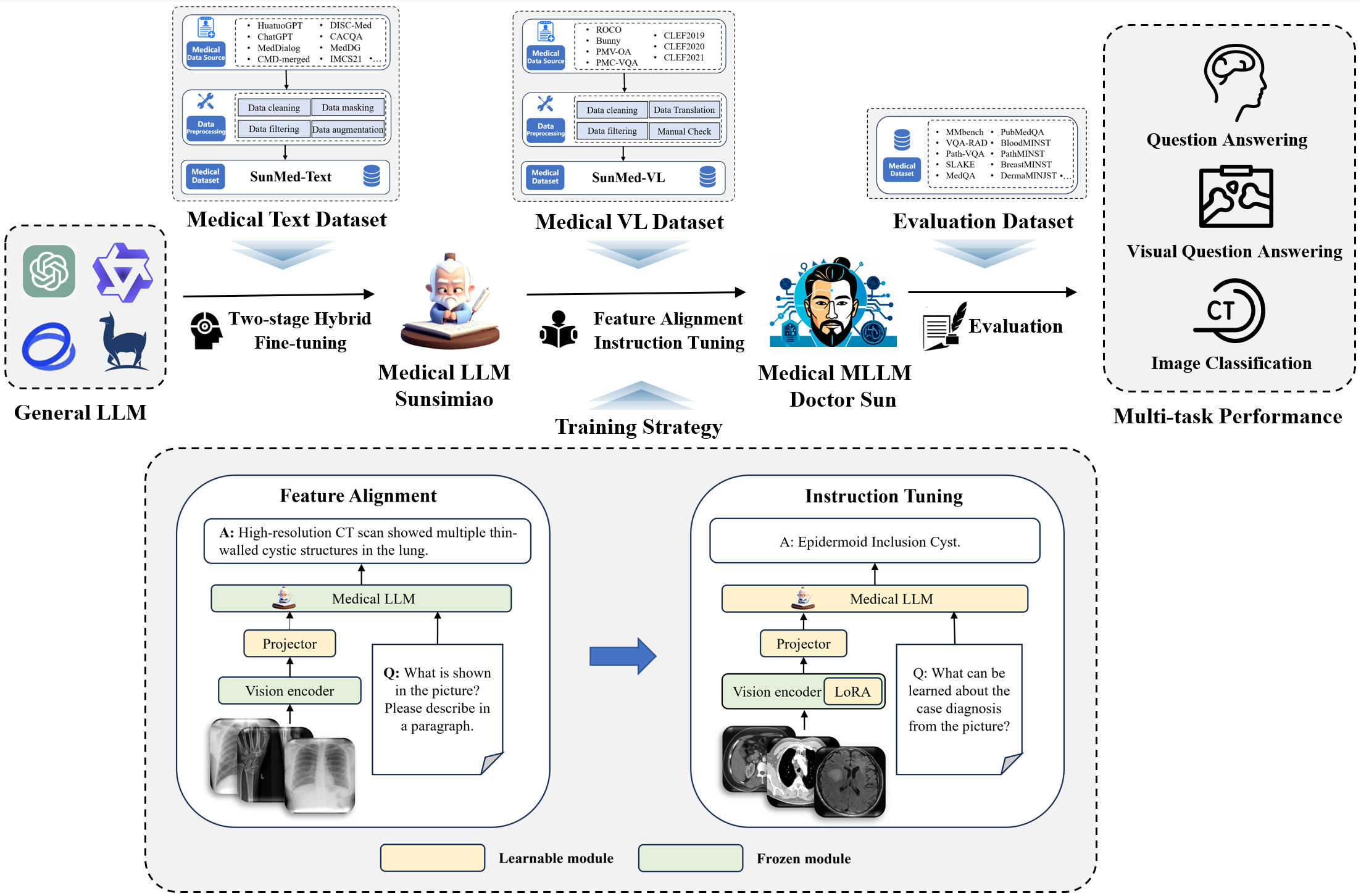}%
}
\caption{~The main architecture and training process of Doctor Sun.}

\label{fig: Overview of Doctor Sun}
\end{figure}

Fig.~\ref{fig: Overview of Doctor Sun} shows the main methods of Doctor Sun. First of all, we collect a large amount of open-source texts and multimodal datasets, and carry out rigorous data processing. Then, the medical LLM is trained and used as the language backbone of Doctor Sun. Then, Doctor Sun is trained through two steps, including feature alignment and instruction fine-tuning. Finally, we conduct evaluations on a wide range of medical datasets to demonstrate the multi-task performance.

\subsection*{Data Collection and Pre-processing}

To ensure the scale and diversity of the data and the ability of the model to perform diverse diagnostics, we collect both text and multimodal large and diverse data for the medical LLM and Doctor Sun.

For the medical LLM, a comprehensive repository of candidate data is constructed by collecting existing medical datasets, with a notable inclusion of a significant number of Chinese medical datasets. From thirty-eight medical datasets, over ten million instructional data points are aggregated. These data points include instructions generated by ChatGPT and real-world instructions from physicians. The content encompasses medical questions and answers, health knowledge, diagnostic procedures, medication recommendations, and other relevant information. In addition to medical-specific data, generic instructions aimed at enhancing doctor-patient interactions are incorporated, thus increasing the utility of the repository in real-world applications. Detailed information about the datasets can be found in the literature~\cite{2024Doctor-Sun}.

To address the variable quality of the data and the presence of duplicates, several procedures are implemented to ensure a high-quality training dataset, ultimately screening approximately one million data points from the original dataset. First, to address potential overlap among different data sources, the MD5 hashing exact matching method is employed to deduplicate text samples and remove repeated sentences. Second, the medical BERT model is used to extract text features, and the distance between features is calculated to exclude approximate data. These steps help mitigate issues related to uneven data distribution and prevent degradation in generalization performance. Additionally, text-matching techniques identify and eliminate data containing personal information,  safeguarding patient privacy. Finally, sentences containing special and non-Chinese characters are removed, and perplexity metrics are used to identify and exclude irrational sentences, ensuring the overall quality of the training data.

For the Doctor Sun, we collect the following datasets: VQA-Med-2019, VQA-Med-2020, VQA-Med-2021, PMC-OA, PMC-VQA, ROCO, datasets of Bunny. For more details, please refer to Table~\ref{Details of data sets}.

Given that the datasets are primarily in English, we preprocess the datasets to enhance the bilingual capability of the model. As an initial step, we select VQA-Med-2019, VQA-Med-2020, VQA-Med-2021, and ROCO for English-Chinese translation. Medical English translation requires a high level of professionalism, so we use GLM-4-Plus as our medical translation assistant, and use it to judge the language fluency of the data. We eliminate the data with poor language fluency to ensure the quality of the dataset. Finally, we manually check the translated data to eliminate any anomalies.

PMC-OA and ROCO are used for feature alignment, which are image description tasks that are naturally suitable for aligning images and text modalities. Additionally, we utilize VQA-Med-2019, VQA-Med-2020, VQA-Med-2021, and PMC-VQA for instruction tuning, which are image question answering tasks, to further enhance the fine-grained understanding and reasoning abilities of the model for images. Finally, the appropriate amount of data is randomly extracted from the dataset of Bunny according to the required mixing ratio and added to the training dataset as generic data. All the selected data comprises SunMed-VL, a selective mixed Chinese-English medical visual-language dataset.

To be clear, the discrepancy between the total number of selected data and the total number of SunMed-VL can be attributed to individual records within the CLEF dataset containing multiple questions and answers, which are expanded into multiple question pairs.

\begin{table}[t]
\caption{Details of multimodal medical datasets}
\centering
    \small

    \begin{tabular}{cccp{8cm}cc}
        \toprule

Dataset & Task & Language &        Description& Number &Selection Ratio\\
        \midrule
        Bunny-Pre~\cite{He2024EfficientML}& VQA& EN& One high-quality, diverse dataset derived from LAION-2B~\cite{schuhmann2022laion} by a three-step corset selection scheme based on CLIP embedding.& 2000000 &12.5\%\\
 PMC-OA~\cite{lin2023pmc}& IC& EN& A large dataset of 12 diagnostic procedures and 12 diseases in the biomedical domain.& 1646592 &12.1\%\\
 ROCO~\cite{pelka2018radiology}& IC& EN&  A medical and multimodal imaging dataset with more than nine medical imaging modalities.& 81000 &61.7\%\\
 Bunny-Fine~\cite{He2024EfficientML}& VQA& EN&A set of visual instruction tuning datasets from SVIT-mix-665K~\cite{zhao2023svit}, WizardLM-evol-instruct-70K~\cite{abbas2023semdedup}.& 695000&18.0\%\\
 VQA-Med-2019~\cite{ImageCLEF-VQA-Med2019}& VQA  & EN   & A unique, dedicated dataset for medical visual question answering, focusing on four key clinical question areas - modality, plane, organ system, and abnormality.& 15292 &32.4\%\\
 VQA-Med-2020~\cite{ImageCLEF-VQA-Med2020}& VQA  & EN   & Visual question generation and medical question-answer task, which covers a variety of conditions.& 5000 &62.0\%\\
 VQA-Med-2021~\cite{ImageCLEF-VQA-Med2021}& VQA  & EN   & Visual question generation and medical question-answer task.& 5500 &57.7\%\\
 PMC-VQA~\cite{zhang2023pmc}& VQA& EN& A medical visual question-answering task covers more than 20 types of image data, including images from radiology and pathology to microscopic imaging and other modalities.& 227000 &100\%\\ 
\midrule
 SunMed-VL& VQA& ZH-EN& A selective mixed Chinese-English medical visual-language dataset covering a variety of diseases and image categories.& 875000&-\\
\bottomrule
    \end{tabular}        
\label{Details of data sets}
\end{table}

\subsection*{Model Architecture}
The proposed Doctor-Sun model comprises three pivotal modules. Firstly, a vision encoder serves as a feature extractor, decoding a sequence of token embeddings from an input image. In the medical domain, data often contains paired multi-modal clinical features, such as text descriptions and corresponding radiological images. Meanwhile, the CLIP model is trained using a pre-training method that compares text-image pairs, enabling it to learn the matching relationship between text and images. Therefore, we utilize a CLIP model to obtain image embeddings.

Additionally, CLIP can effectively Utilize multimodal medical data and achieve high classification accuracy even with limited labeled data. This capability is particularly crucial in the medical field, where data sparsity is a significant challenge. Besides, LORA adapters are added to the CLIP vision encoder to enable the vision model to learn more fine-grained medical image features and medical concepts in instruction tuning. Moreover, a trainable projection module, consisting of a single linear layer, transforms the image features into the language embedding space, seamlessly bridging the dimensional gap between the visual and linguistic domains. Lastly, we utilize our trained medical LLM as the linguistic backbone. In essence, for a given input question $Q$, the textual content is converted into a corresponding word embedding, then synchronized with the aligned visual features as input for the LLM. The model predicts the subsequent embedding based on the preceding one, iterating the process until the desired result is achieved. 

\subsection*{Model Training}
This subsection introduces the medical language backbone training process, including the two-stage hybrid fine-tuning and the Doctor Sun training process, which involves aligning biomedical image-text features and instruction tuning.

\subsubsection*{Two-stage Hybrid Fine-tuning}
The medical language backbone builds upon llama3.1-8b-instruct-dpo-zht~\cite{llama3.1-8b-instruct-dpo-zh}, a LLM incorporating instruction tuning and direct preference alignment.

Fine-tuning LLMs solely with medical data introduces a high risk of catastrophic forgetting due to the substantial divergence between medical and general datasets. This process often diminishes the model's general reasoning abilities, which are essential for solving complex problems. During fine-tuning, parameter weights become increasingly specialized to the medical domain, analogous to the loss of previously acquired knowledge when focusing exclusively on a single discipline without reinforcement.

Using a mixed dataset of medical and general data for fine-tuning also leads to inefficiencies. Training on such mixtures increases the volume of general data alongside domain-specific data, resulting in unnecessary computational overhead. Adjusting the proportion of domain-specific data in these mixtures has minimal impact on general capabilities when the general data volume is fixed~\cite{dong2023abilities}. Consequently, single-stage mixed-data fine-tuning is computationally suboptimal.

To address these challenges, a two-stage hybrid fine-tuning strategy is adopted. The first stage employs approximately 300,000 medical data points for supervised fine-tuning, equipping the model with detailed domain-specific knowledge and enhancing its diagnostic proficiency. The second stage fine-tunes the model using a dataset comprising $80\%$ general data and $20\%$ medical data, totaling 250,000 instances. This approach balances specialization and generalization, preserving the general reasoning capabilities of models while maintaining their expertise in the medical domain.

\subsubsection*{Biomedical Image-Text Feature Alignment}
As a multimodal LLM, Doctor Sun must understand and respond to user instructions expressed as natural language queries, containing information from the visual modality. To align the modalities between vision and text, the training process begins with the image captioning task. During the training of feature alignment, the visual encoder and LLM remain frozen, and only the projector is trained. The training data is represented as $\text{D}=\{{{({{X}_{v}}, {{X}_{t}})}^{1}}, \ldots, {{({{X}_{v}}, {{X}_{t}})}^{n}}\}$, where $n$ is the total number of data points. Each sample includes an image ${{X}_{v}}$ and a description ${{X}_{t}}$, and each pairing of image and text inputs is converted into corresponding sequence embeddings (${{H}_{v}}$ and ${{H}_{t}}$). The negative log-likelihood objective function in the auto-regressive model is calculated as follows:
\begin{equation}
\sum\limits_{v,t\in D}{-\log p({{H}_{t}}\left| {{H}_{v}} \right.)}=\sum\limits_{v,t\in D}{-\log \prod\limits_{i=1}^{L}{p(H_{t}^{i+1}| {{H}_{v}}, H_{t}^{(1:i)})}}
\end{equation}
where $L$ denotes the token length in ${{H}_{t}}$ and $H_{t}^{(1:i)}$ represents all the embeddings before $H_{t}^{i+1}$.

This initial training step aligns the different modalities and accumulates knowledge of medical-specific terms and images.

\subsubsection*{Instruction Tuning}
To enhance the ability of the model to follow instructions and perform disease diagnosis, the second step of instruction tuning is employed. In the current method~\cite{li2024llava}, researchers focus solely on fine-tuning the modal projector and the LLM during instruction tuning. Consequently, the general vision module struggles to extract disease features from medical images due to a lack of medical knowledge and the significant disparity between medical images and ordinary ones. Therefore, based on the previous multimodal understanding method~\cite{liu2024visual}, the visual encoder, the modal projector, and the LLM are unfrozen, allowing the visual encoder to learn medical knowledge gradually.

A training strategy incorporates LORA tuning on the CLIP vision encoder and full tuning on the LLM. Unfreezing the pre-trained backbones under the fully autoregressive architecture can lead to training divergences, and leveraging LORA can enhance expressivity and stabilize the training process~\cite{laurenccon2024matters}. However, considering that the performance of LORA fine-tuning is often close to or worse than full fine-tuning~\cite{sun2023comparative}, and given that the medical LLM already possesses substantial medical knowledge, a hybrid fine-tuning approach is applied.

The training data is processed into $\text{D}=\{{{({{X}_{v}}, {{X}_{q}}, {{X}_{a}})}^{1}}, \ldots, {{({{X}_{v}}, {{X}_{q}}, {{X}_{a}})}^{n}}\}$, where $n$ represents the total number of data points. Each sample consists of an image ${{X}_{v}}$, a question ${{X}_{q}}$, and an answer ${{X}_{a}}$. Each sample is then mapped to a sequence embedding (${{H}_{v}}$, ${{H}_{q}}$, ${{H}_{a}}$). During training, the negative log-likelihood loss is minimized as follows
\begin{equation}
\mathsf{Loss}=-\log \prod\limits_{v,q,h\in D}^{{}}{p(H_{a}^{i+1} | {{H}_{v}}, H_{q}^{(i+1)}, H_{h}^{(1:i)})}
\end{equation}
where $i$ represents the number of dialogue rounds in a set of conversations in dataset $D$, and $H_{h}^{(1:i)}$ represents the history before the current conversation in a set of conversations.

\subsection*{Evaluation Datasets}
Any evaluation primarily focuses on assessing the responsiveness of models to a wide range of question prompts, thereby defining the scope of tasks the model is proficient in executing. Assessments are conducted across two tasks to evaluate the capabilities of the MLLM: question answering (QA) and visual question answering (VQA). Detailed information is shown in Table~\ref{Details of evaluation datasets}.

\textbf{Question Answering.} To evaluate the ability to understand and reason about scientific biomedical literature and the mastery of medical knowledge, the model is assessed using two datasets: MedQA features multiple-choice questions, while PubMedQA is designed with closed-ended questions. For MedQA, the model receives a question and all the options provided to determine the correct answer. For PubMedQA, the model must answer questions in the form of yes, no, or maybe. Answering performance is reported based on accuracy.

\begin{table}[t] 
    \caption{Details of evaluation datasets}
    
    \label{Details of evaluation datasets}  

    \begin{tabular*}{\linewidth}{@{\extracolsep{\fill}}cccc}
        \toprule
        Dataset & Task& Modality& Volume
 \\
\midrule
 MMbench-dev~\cite{MMBench}& VQA& General&4329 \\
 VQA-RAD~\cite{lau2018dataset}& VQA& Radiology&451 \\
         Path-VQA~\cite{He2020PathVQA3Q}& VQA
& Radiology
& 1061
 \\
 SLAKE~\cite{liu2021slake}& VQA
& Pathology
&6719
 \\
 MedQA~\cite{jin2021disease}& QA
& General Medicine
&1273
 \\
 PubMedQA~\cite{jin2019pubmedqa}& QA
& General Medicine
&500
 \\
\bottomrule

\end{tabular*}
\end{table}

\textbf{Visual Question Answering.} To evaluate the proficiency of the model in medical visual question answering, three datasets are utilized, encompassing a blend of open-ended and closed-ended inquiries: Path-VQA, SLAKE, and VQA-Rad. Additionally, MMbench is used to assess the multimodal understanding capability of large vision language models. Due to constraints on the number of uploadable answers for verification and the need for high evaluation frequency, the development set from MMBench is selected for assessment to facilitate rapid local evaluations. To accurately evaluate the Chinese ability of the model, Doctor Sun used Chinese for the MMbench-dev evaluation, while LLaVA-Med and RadFM used English to avoid the problem of inaccurate evaluation caused by the language gap. The VQA prompt is created by combining the image with its related question. Separate metrics are reported for closed-ended and open-ended questions, including closed-ended accuracy, open-ended accuracy, and open-ended recall. Furthermore, overall recall, F1 score, and BLEU-1 score are provided to evaluate the model from a broader perspective.

MultiMedEval is the medical benchmark evaluation tool that ensures a standardized and objective evaluation based on established methodologies~\cite{royer2024multimedeval}. Additionally, the built-in tools from Xtuner~\cite{2023xtuner} are utilized to assess the performance of MMBench-dev.

To ensure accuracy, the test dataset is excluded from all training steps. Additionally, due to the lack of current Chinese multimodal medical benchmark evaluations, Doctor Sun conducted the biomedical evaluation entirely in English. Finally, the comprehensive ability assessment conducted by Doctor Sun is in Chinese, while the comparison models are evaluated in English. We utilize a range of evaluation metrics to assess the capabilities of our Doctor Sun model across various tasks. Accuracy and recall rate are the main metrics to evaluate performance in VQA and QA tasks, including open question accuracy (O-A), closed question accuracy (C-A), open question recall (O-R), and overall recall. Given the issue of category imbalance in medical VQA tasks, we employ the F1 score for evaluation. The F1 score represents the harmonic mean of precision and recall. To evaluate the lexical match with standard 

\section*{Results}
In this section, we conduct experiments to explore three key questions: (1) Is the use of specialized LLMs more effective than general-purpose LLMs for domain-specific multimodal tasks? (2) What is the optimal ratio of domain-specific data to generic data for feature alignment and instruction tuning, ensuring the model retains general and domain-specific capabilities? (3) How does the performance of Doctor Sun compare to current biomedical MLLMs under standard benchmarks?

\subsection*{Performance on Established Benchmarks}

\begin{table}[t]
\caption{Multimodal medical benchmark results}
\label{Performance on Established Benchmarks}
\centering
 
\begin{tabular*}{\linewidth}{@{\extracolsep{\fill}}ccccccccc}
\toprule
Dataset& Metric&       RadFM&           LLaVA-Med & LLaVA-Med-v1.5&          DocS-M&             DocS-GL&DocS-FV\\ 
\hline
\multirow{6}{*}{VQA-Rad}      & BLEU-1            & \underline{0.475}& 0.033      & 0.064& \textbf{0.490}& 0.463
 &0.346
\\
                              & Closed Q accuracy & 0.577& 0.545      & 0.558& \textbf{0.641}& \underline{0.622}&0.562
\\
                              & open Q recall& \textbf{0.407}& 0.246      & \underline{0.356}& 0.339& 0.317
 &0.181
\\
 & recall            & 0.438& 0.372      & 0.468& \textbf{0.508}&\underline{0.486}&0.393
\\
                              & open Q accuracy   & \textbf{0.335}& 0.140      & 0.224& \underline{0.255}& 0.240&0.100\\
                              & F1                & 0.442& 0.069      & 0.110& \textbf{0.501}& \underline{0.476}&0.357
\\ 
\hline
\multirow{6}{*}{Slake-VQA}    & BLEU-1            & \textbf{0.746}& 0.036      & 0.060& \underline{0.417}& 0.412
 &0.098
\\
                              & Closed Q accuracy & \textbf{0.752}& 0.512      & 0.566& \underline{0.617}& 0.538
 &0.459
\\
                              & open Q recall     & \textbf{0.758}& 0.429& \underline{0.468}& 0.378& 0.374
 &0.239
\\
                              & recall            & \textbf{0.695}& 0.443      & \underline{0.500}& 0.458& 0.441
 &0.312
\\
                              & open Q accuracy   & \textbf{0.725}& 0.362& \underline{0.408}& 0.334& 0.324
 &0.183
\\
                              & F1                & \textbf{0.714}& 0.075      & 0.107& \underline{0.429}& 0.422
 &0.130\\ 
\hline
\multirow{6}{*}{Path-VQA}     & BLEU-1            & 0.257& 0.021      & 0.039& \textbf{0.304}& \underline{0.302}&0.297
\\
                              & Closed Q accuracy & 0.505 & 0.512      & \underline{0.570}& \textbf{0.571}& 0.557&0.553
\\
                              & open Q recall~    & 0.020 & \textbf{0.116} & \underline{0.090}& 0.066& 0.061
 &0.038
\\
                              & recall            & 0.221 & 0.287      & \underline{0.317}& \textbf{0.320}& 0.309&0.308
\\
                              & open Q accuracy   & 0.005 & \textbf{0.053} & 0.045& \underline{0.046}& 0.044
 &0.009
\\
                              & F1                & 0.232& 0.052       & 0.071& \textbf{0.310}& \underline{0.307}&0.300\\ 
\hline
MedQA                         & Accuracy          & 0.230 & 0.241      & 0.338& \underline{0.353}& 0.349
 &\textbf{0.368}\\ 
\hline
PubMedQA                      & Accuracy          & 0.336 & 0.488      & 0.490& \underline{0.736}& 
0.698
 &\textbf{0.764}\\
\hline
\end{tabular*}
\end{table}

Doctor Sun (DocS-M) is evaluated against RadFM, which is an open-source implementation of a medical MLLM~\cite{wu2023towards}, LLaVA-Med, and LLaVA-Med-v1.5-mistral-7b, multimodal medical models trained on hundreds of thousands of medical data using the LLaVA framework~\cite{li2024llava}. The necessity of unfreezing the visual encoder and employing a medical-specific LLM is also demonstrated.

To ensure realistic and reproducible results, Doctor Sun is tested without fine-tuning on downstream tasks. All evaluations are conducted using MultiMedEval~\cite{royer2024multimedeval}, and the results are summarized in Table~\ref{Performance on Established Benchmarks}.

\begin{figure}[t]

\centering
\subfloat[\centering]{\includegraphics[width=7.0cm]{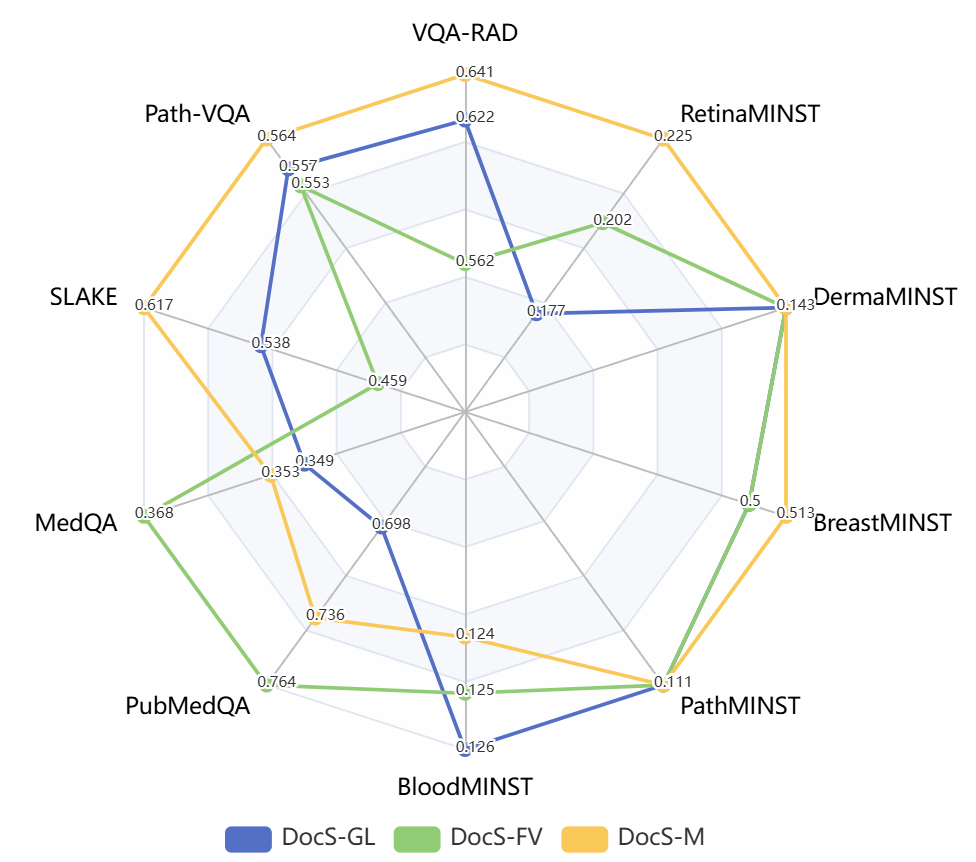}}
\subfloat[\centering]{\includegraphics[width=7.0cm]{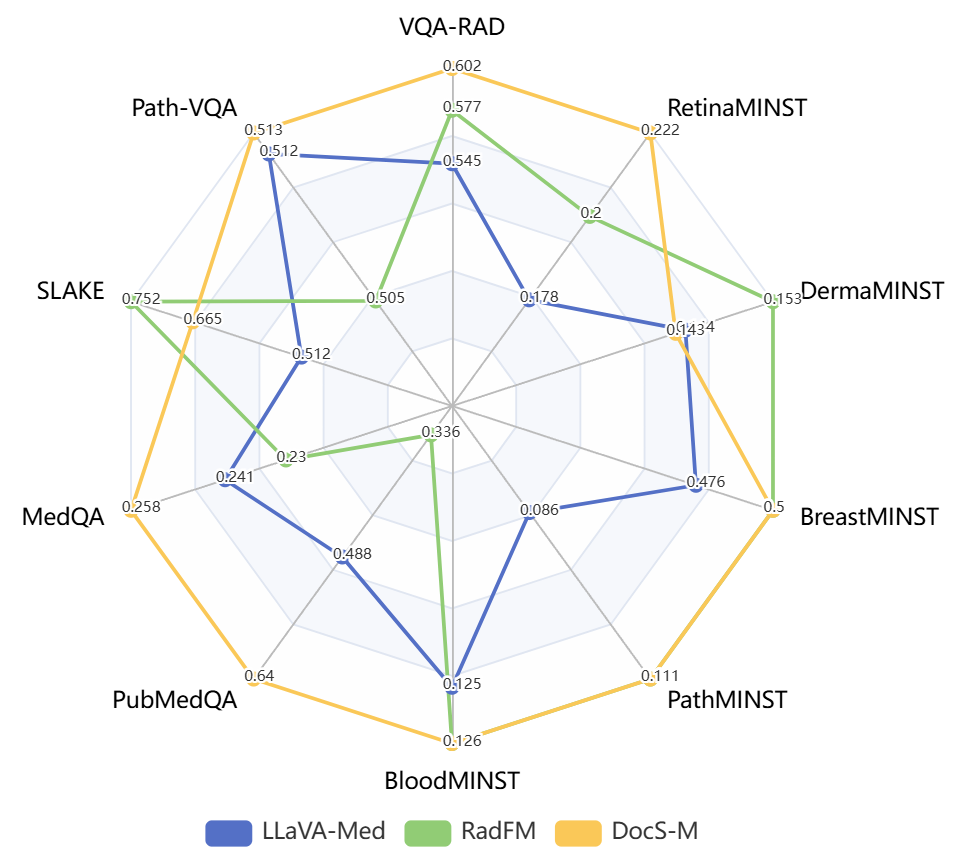}}

\caption{(\textbf{a}) The performance of Doctor Sun under different linguistic backbones. (\textbf{b}) Closed-set performance comparative analysis between DocS-M and related work. \label{figure 4}}
\end{figure}

The performance of different linguistic backbones and visual encoder training strategies is compared. Fig.~\ref{figure 4}a illustrates the average scores across all metrics on various test datasets. In QA and VQA tasks, which heavily rely on the LLM's capabilities, DocS-M (medical LLM with an unfrozen vision encoder) surpasses DocS-GL (general LLM with an unfrozen vision encoder) by $4\%$, highlighting the importance of domain-specific knowledge integration for medical applications. Regarding visual encoder training strategies, DocS-M outperforms DocS-FV (medical LLM with a frozen vision encoder) by $26\%$ in tasks requiring image understanding, such as VQA. This result highlights the limitations of universal vision encoders in extracting features from medical images. However, in QA tasks, DocS-M performs slightly worse than DocS-FV, likely due to the universal encoder's insufficient image comprehension, which forces the LLM to compensate by over-relying on textual medical knowledge.

A detailed comparison of DocS-M with RadFM, LLaVA-Med, and LLaVA-Med-v1.5-mistral-7b is conducted:

\begin{itemize}
    \item \textbf{Question Answering.} QA tasks evaluate the foundational medical knowledge of the models and their ability to apply it across diverse biomedical domains. DocS-M demonstrates significant improvements over LLaVA-Med, LLaVA-Med-v1.5-mistral-7b, and RadFM, with average performance gains of $49\%$, $32\%$, and $92\%$, respectively, indicating a deeper integration of medical expertise.

    \item \textbf{Visual Question Answering.} Medical VQA assesses the ability of models to integrate visual and textual information in a clinical context. DocS-M achieves $62\%$ and $51\%$ performance advantages over LLaVA-Med and LLaVA-Med-v1.5-mistral-7b across almost all evaluation benchmarks. Although RadFM slightly outperforms DocS-M, it benefits from fine-tuning on Rad-VQA and SLAKE datasets, whereas DocS-M operates without such task-specific optimization. Notably, DocS-M achieves a $2\%$ lead over RadFM on the Rad-VQA dataset. On the Path-VQA benchmark, which provides a more balanced comparison, DocS-M surpasses RadFM by $30\%$ while utilizing only $64\%$ of RadFM's parameters. The ability to achieve superior performance with reduced computational resources is particularly advantageous for practical deployment.

\end{itemize}

As illustrated in Fig.~\ref{figure 4}b, DocS-M demonstrates superior performance on nearly all closed-set problems, underscoring its advanced inferential and perceptual capabilities as critical factors for success.

\subsection*{Experiments on General and Specialized Abilities}

To balance general and professional abilities, we train eight model variants with different mixture ratios of training datasets. The final results are presented in Table~\ref{Results of different data mixing ratios}, where $E1$ and $E2$ represent the domain and generic data mixing ratios of $1:0$ and $1:1$ in feature alignment respectively. $V1$, $V2$, $V3$, $V4$ represent the domain and general data mixing ratio of $1:0$, $1:0.2$, $1:0.5$, and $1:1$ in instruction tuning. LLa-M stands for LLaVA-Med, and Doc-S stands for Doctor Sun.

Given the huge time and computational cost of evaluating all models on all benchmarks, and VQA represents a broader and generic challenge in image analysis across all tasks, we compare models on three VQA tasks in this experiment, including two medical benchmarks and one general benchmark. All models use a fixed amount of domain data during training, including a quarter of a million feature alignment data and sixty thousand instruction tuning data.

\begin{figure*}
    \centering
    \includegraphics[width=0.9\linewidth]{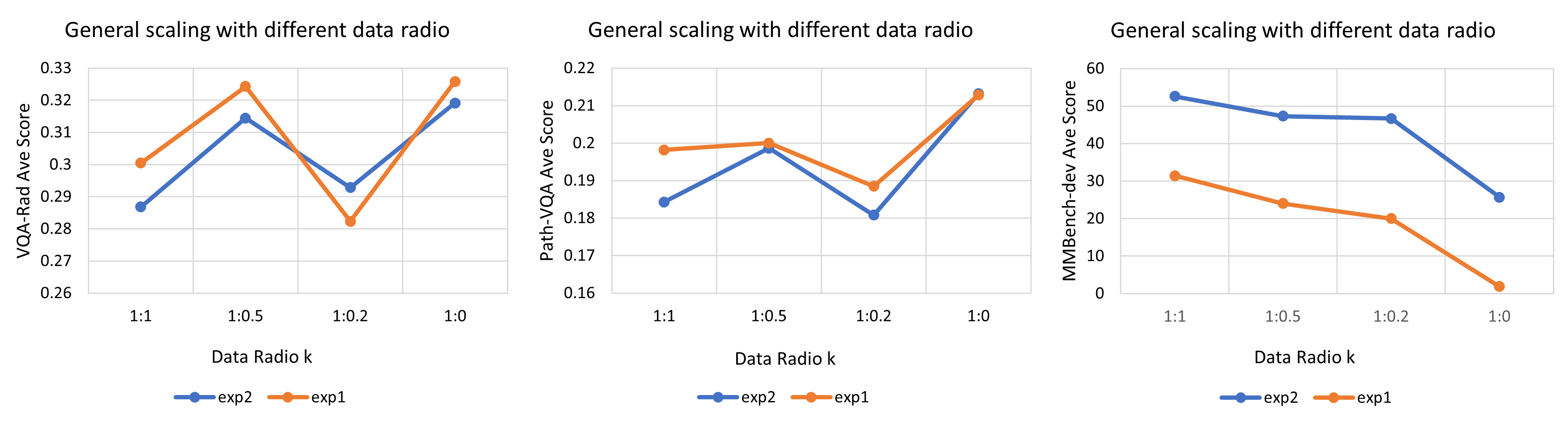}
    \caption{~Different data ratios (k) between specific abilities and general abilities on three benchmarks.}
    \label{figure 3}
\end{figure*}

More specifically, the domain ability of the models is evaluated in various situations. During the feature alignment phase, the model trained exclusively on medical data outperforms the model trained on mixed data by approximately $2.3\%$. In the instruction tuning stage, the model with a $1:0.5$ mixture of domain-specific and generic data excels in visual question answering, achieving the best performance on eleven out of twenty metrics. As shown in Fig.~\ref{figure 3}, models utilizing medical datasets tend to achieve higher average scores, indicating more specialized and concise answers, as evidenced by increased F1 and BLEU-1 metrics. However, despite the professional performance, these models do not excel in recall rate, which is crucial in the medical field since missed diagnoses are more severe than misdiagnoses.

\begin{table}[t]
\small
    \caption{Evaluation results of different data mixing ratios}
    
    \label{Results of different data mixing ratios}
\centering

    \begin{tabular*}{\linewidth}{@{\extracolsep{\fill}}cccccccccccccc}
        \toprule
        &  \multicolumn{6}{c}{VQA-RAD}&  \multicolumn{6}{c}{Path-VQA}& MMbench\\
 Version& O-A& O-R& C-A&  Recall&BLEU-1& F1& O-A& O-R& C-A&   Recall&BLEU-1&F1&Ave\\
        \midrule
        E1-V1& 0.140& 0.200& 0.514
&  0.374
 &\textbf{0.359}& \textbf{0.368}& 0.021& 0.033& 0.474
&   0.253
 &\textbf{0.246}&\textbf{0.250}
&1.9\\
        E1-V2& 0.165& 0.230& 0.506
&  0.383
 &0.182
& 0.228
& 0.025& 0.035& 0.529
&   0.253
 &0.128
&0.161
&20.0
\\
        E1-V3& \textbf{0.185}& \textbf{0.282}& 0.545&  \textbf{0.428} &0.228
& 0.277& \textbf{0.033}& \textbf{0.047}& \textbf{0.530}
&   \textbf{0.288} &0.134
&0.168&24.0
\\
        E1-V4& 0.150& 0.215& \textbf{0.570}&  0.413 &0.203
& 0.252& 0.028& 0.040& 0.515
&   0.277 &0.151
&0.178&\textbf{31.4}
\\
        \midrule
        E2-V1& 0.175& 0.248& 0.471
&  0.396
 &\textbf{0.375}& 0.250
& 0.024& 0.035& 0.471
&   0.253
 &\textbf{0.246}&\textbf{0.250}
&25.7
\\
        E2-V2& 0.150& 0.219& \textbf{0.554}
&  \textbf{0.406}
 &0.188
& 0.240
& 0.024& 0.033& 0.503
&   0.268
 &0.112
&0.145
&46.7
\\
        E2-V3& \textbf{0.195}& \textbf{0.265}& 0.494
&  0.392
 &0.253
& \textbf{0.288}
& 0.023& 0.034& \textbf{0.537}
&   0.270
 &0.149
&0.179
&47.3
\\
        E2-V4& 0.170& 0.247& 0.498
&  0.387
 &0.187
& 0.232
& \textbf{0.030}& \textbf{0.042}& 0.512&   \textbf{0.278}
 &0.103
&0.141
&\textbf{52.6}
\\
 \midrule
 Doc-S& 0.255& 0.339& \textbf{0.641}&  \textbf{0.508}&\textbf{0.490}& \textbf{0.501}& 0.046& 0.066& \textbf{0.571}&   \textbf{0.320}&\textbf{0.304}&\textbf{0.310}& \textbf{53.6}\\
 LLa-M& 0.140& 0.246& 0.545& 0.372 &0.033& 0.069& \textbf{0.053}& \textbf{0.116}& 0.512& 0.287 &0.021& 0.052&0.3\\
 RadFM& \textbf{0.335}& \textbf{0.407}& 0.577&  0.438 &0.475& 0.442& 0.005& 0.020& 0.505&   0.221 &0.275&0.232& 0.1\\
 \bottomrule

    \end{tabular*}

\end{table}

Moreover, the generalizability of the models is also assessed. In the feature alignment stage, models employing data mixing demonstrated a significant advantage, showing an average improvement of $123.3\%$ compared to those that did not. In the instruction tuning stage, models with greater amounts of training data generalizability consistently exhibit superior performance. Several existing works~\cite{wu2023towards,li2024llava} are evaluated and showed poor performance, with most answers being either empty or irrelevant to the questions, resulting in low generalization performance. Manual testing confirmed that while the models correctly answer medical questions, the ability to follow general instructions is lacking, precluding realistic doctor-patient interactions.

Experimental results indicate that models experience catastrophic forgetting of general abilities when relying solely on domain data during two-stage training, including general image captioning proficiency and perception and reasoning capabilities. Incorporating general training data during the feature alignment phase significantly improved general ability by over $100\%$ while maintaining comparable domain ability. In the instruction tuning stage, increasing the proportion of general data enhanced overall ability. Integrating an appropriate amount of generic data during training positively influenced the diagnostic performance of the model, attributed to improved reasoning and perception capabilities.

To further balance the specialized and general capabilities while minimizing training costs without compromising performance, data mixing ratios of $1:1$ and $1:0.5$ are respectively chosen for feature alignment and instruction tuning.

\section*{Discussion}
Doctor Sun demonstrates outstanding zero-shot diagnostic capabilities in biomedical tasks, particularly in VQA and QA tasks. Our research results highlight the importance of integrating domain-specific knowledge into the foundational components (visual encoder and language model) of multimodal large models during domain-specific fine-tuning, as well as the necessity of combining general data with domain-specific data to balance the general and professional abilities. At the same time, the addition of bilingual capabilities enables Doctor Sun to effectively integrate into actual clinical workflows, especially in multilingual medical environments. 
Additionally, to deeply explore the effectiveness of the language backbone of Doctor Sun, we also conduct specialized evaluations and analyses of its core medical LLM.

To evaluate the medical capabilities of medical LLM, a range of medical LLMs are selected as benchmarks for comparative analysis, including DISC-MedLLM-13B, HuatuoGPT-7B, Mixtral-8x7B, and Bentsao-7B~\cite{Bao2023DISCMedLLMBG, zhang2023huatuogpt, jiang2024mixtral, Wang2023HuaTuoTL}. Besides, CMB is used as a medical evaluation dataset~\cite{wang2023cmb}, which is designed to assess the depth and breadth of medical knowledge mastered by medical models across various aspects and levels of difficulty. Additionally, to fully evaluate the impact of two-stage hybrid fine-tuning on the general abilities, BIG-Bench Hard (BBH)~\cite{suzgun2022challenging}, Mostly Basic Python Problems (MBPP)~\cite{austin2021program}, and C-Eval~\cite{huang2024c} are also utilized to evaluate the reasoning ability, Python programming ability, and Chinese multi-level multidisciplinary knowledge. We use the Chinese version for CMB and the English version for the other benchmarks to demonstrate bilingual capabilities. This comprehensive evaluation allows us to assess the impact of two-stage hybrid fine-tuning on the general and cross-domain capabilities of the model, demonstrating the effectiveness of the training strategy.

\begin{table}[t!]
\centering 
\begin{minipage}[t]{0.55\textwidth}
\makeatletter\def\@captype{table}
\centering
\caption{Evaluation results different training stages}
\label{table:training_stages}
    \begin{tabular}{ccccc}
    \toprule
    Training stage & BBH   & C-Eval & MBPP  & CMB   \\
    \midrule
    Qwen-14B       & 0.537 & 0.717  & 0.398 & -     \\
    InternLM-20B   & 0.525 & 0.588  & 0.356 & -     \\
    Qwen1.5-7B     & 0.402 & 0.741  & 0.516 & -     \\
    \midrule
    stage1         & 0.401 & 0.500  & 0.528 & 0.371 \\
    stage2         & 0.559 & 0.518  & 0.538 & 0.415 \\
    \bottomrule
    \end{tabular}
\end{minipage}
\hfill 
\begin{minipage}[t]{0.4\textwidth}
\makeatletter\def\@captype{table}
\centering
\caption{Evaluation results on cmb}
\label{table:cmb_results}
    \begin{tabular}{ccc}
    \toprule
    Model         & CMB   & Volume \\
    \midrule
    DISC-MedLLM   & 0.398 & 50w       \\
    Mixtral-8x7B  & 0.363 & -         \\
    HuatuoGPT     & 0.320 & 1w        \\
    Bentsao       & 0.204 & 1w        \\
    \midrule
    Doctor Sun    & 0.415 & 35w       \\
    \bottomrule
    \end{tabular}
\end{minipage}
\end{table}

To investigate the impact of two-stage training on general and specialized capabilities, we first report the accuracy of all evaluation datasets during both stages in Table~\ref{table:training_stages}. The second stage of training yields comprehensive improvements, particularly in reasoning ability, which is central to human general intelligence. The BBH score is enhanced by $39\%$, and it also has significant advantages over general models such as Qwen-14B~\cite{bai2023qwen}, InternLM-20B~\cite{team2023internlm}, and Qwen1.5-7B~\cite{qwen1.5github}, demonstrating the effectiveness and necessity of the two-stage hybrid training approach. Additionally, the performances of the two-stage models on the C-EVAL and MBPP benchmarks are comparable, suggesting that knowledge acquisition in these models is an accumulative rather than a superimposed process. The large gap on the C-EVAL is due to the gap in knowledge accumulation, where we care more about reasoning ability than general knowledge. Additionally, we compare similar works on the medical evaluation benchmark CMD in Table~\ref{table:cmb_results}. The medical LLM we trained possesses a degree of proficiency within the medical domain. The performance improvement on the CMB benchmark can be attributed to knowledge accumulation and enhanced reasoning ability. Our evaluation results on three datasets demonstrate the efficacy of two-stage training in enhancing reasoning abilities, showing that the general ability and domain-specific ability complement each other.

Despite these advancements, the study has some limitations. Doctor Sun has not been evaluated in a real-world environment, and while its performance on medical evaluation benchmarks is commendable, it is not yet suitable for direct clinical application. Future work will involve more extensive testing to determine the impact of integrating domain-specific and generic data on the performance of multimodal models in other specialized domains, aiming to draw more generalized conclusions.

\section*{Conclusion}

In summary, this pioneering research introduces an innovative Chinese-English bilingual multimodal medical diagnostic assistant named Doctor Sun, along with a meticulously curated mixed Chinese-English medical visual-language dataset, SunMed-VL. These advancements are rigorously tested and validated through extensive experiments, revealing that Doctor Sun possesses exceptional zero-shot multimodal medical diagnosis capabilities, surpassing the current medical models across multiple key performance indicators. By integrating medical knowledge into the visual encoder and LLM, Doctor Sun ensures more reliable and accurate medical diagnoses, improving diagnostic accuracy and reducing the risk of error. Additionally, incorporating an appropriate mix of domain-specific medical and general-purpose data enhances professional proficiency and general versatility, improving its effectiveness in various real-world medical application scenarios.

\section*{Acknowledgments}

The authors thank Xin Wang for his assistance in the download and cleaning of some datasets.

\section*{Data availability}
The datasets used and/or analyzed in the current study are available at \url{https://github.com/X-D-Lab/Doctor-Sun}.

\bibliographystyle{unsrt} 
\bibliography{sample.bib}

\end{document}